\documentclass[sn-mathphys,Numbered]{sn-jnl}

\usepackage{graphicx}%
\usepackage{multirow}%
\usepackage{amsmath,amssymb,amsfonts}%
\usepackage{amsthm}%
\usepackage{mathrsfs}%
\usepackage[title]{appendix}%
\usepackage{xcolor}%
\usepackage{textcomp}%
\usepackage{manyfoot}%
\usepackage{booktabs}%
\usepackage{algorithm}%
\usepackage{algorithmicx}%
\usepackage{algpseudocode}%
\usepackage{listings}%
\usepackage{subcaption}

\theoremstyle{thmstyleone}%
\theoremstyle{thmstyletwo}%

\theoremstyle{thmstylethree}%

\begin{document}

\title[Article Title]{Article Title}


\author[1]{\fnm{Zhewei} \sur{Liu}}\email{zheweiliu@tamu.edu}

\author[1]{\fnm{Lipai} \sur{Huang}}\email{lipai.huang14@gmail.com}

\author*[2]{\fnm{Chao} \sur{Fan}}\email{cfan@g.clemson.edu}

\author[1]{\fnm{Ali} \sur{Mostafavi}}\email{mostafavi@tamu.edu}

\affil*[1]{\orgdiv{UrbanResilience.AI Lab, Zachry Department of Civil and Environmental Engineering}, \orgname{Texas A\&M University}, \orgaddress{\city{College Station}, \postcode{77843}, \state{Texas}, \country{U.S.}}}

\affil[2]{\orgdiv{School of Civil and Environmental Engineering and Earth Sciences}, \orgname{Clemson University}, \orgaddress{ \city{Clemson}, \postcode{29634}, \state{South Carolina}, \country{U.S.}}}



\abstract{Generating realistic human flows across regions is essential for our understanding of urban structures and population activity patterns, enabling important applications in the fields of urban planning and management. However, a notable shortcoming of most existing mobility generation methodologies is neglect of prediction fairness, which can result in underestimation of mobility flows across regions with vulnerable population groups, potentially resulting in inequitable resource distribution and infrastructure development. To overcome this limitation, our study presents a novel, fairness-aware deep learning model, FairMobi-Net, for inter-region human flow prediction. The FairMobi-Net model uniquely incorporates fairness loss into the loss function and employs a hybrid approach, merging binary classification and numerical regression techniques for human flow prediction. We validate the FairMobi-Net model using comprehensive human mobility datasets from four U.S. cities, predicting human flow at the census-tract level. Our findings reveal that the FairMobi-Net model outperforms state-of-the-art models (such as the DeepGravity model) in producing more accurate and equitable human flow predictions across a variety of region pairs, regardless of regional income differences. The model maintains a high degree of accuracy consistently across diverse regions, addressing the previous fairness concern. Further analysis of feature importance elucidates the impact of physical distances and road network structures on human flows across regions. With fairness as its touchstone, the model and results provide researchers and practitioners across the fields of urban sciences, transportation engineering, and computing with an effective tool for accurate generation of human mobility flows across regions.}



\title{FairMobi-Net: A Fairness-aware Deep Learning Model for Urban Mobility Flow Generation}

\maketitle
\section{Introduction}\label{sec1}

Assessing human mobility is crucial to urban studies and planning as it provides
valuable insights into how people move within urban areas\cite{wang2022zooming,chen2019stlp,shen2022novel}. Human mobility analysis captures everyday commuting patterns, periodic travels, and exceptional movements during events or emergencies \cite{fan2021fine,liu2022categorisation,rajput2023latent}. Understanding these dynamics enables urban planners to design infrastructure, public transportation systems, and public spaces to meet the needs of the population efficiently and sustainably. Furthermore, by considering mobility patterns, planners can enhance the livability of cities, reduce congestion, improve air quality, and promote social equity \cite{kitamura1988evaluation,liu2021analysis,li2023mobility}. The assessment of human mobility also plays a vital role in urban resilience, helping cities prepare for and adapt to challenges, such as climate change, demographic shifts, or pandemic outbreaks \cite{wang2020review,huang2022social,liu2023beyond}. Human mobility generation and prediction, an emergent field of study fueled by the ubiquity of mobile phones and location acquisition technologies, is fundamentally changing the way we comprehend human behavior and dynamics in urban settings. These technologies have enabled examination of people's mobility patterns, such as the places frequented and their visitation patterns, which shed light on lifestyles \cite{Zion2018IdentifyingAP,fan2022human,huang2023emergence}. Importantly, these lifestyles can act as a predictive tool for individual and collective behavior, a facet extensively explored in marketing, transportation, health, psychology, sociology, among other fields of study. \cite{https://doi.org/10.1111/j.1467-954X.1983.tb00387.x, /content/paper/5kmfp51f5f9t-en,https://doi.org/10.1111/j.1475-097X.1981.tb00894.x,fan2021evaluating}.\\
\indent Human mobility generation and prediction enable evaluation of  future changes in population flows and movements as cities grow and evolve. In addition to prediction of future changes in human flow patterns, human mobility prediction and generation can provide reliable estimates of current human mobility flows when human mobility data are not obtainable due to privacy or availability issues \cite{liu2019recommending,shi2021regnet,coleman2022human}. A plethora of methodologies have been employed in the realm of human mobility prediction, each with its unique advantages and applications. From mathematical models, such as gravity models and radiation models \cite{simini2012universal}; to machine learning (ML)techniques, including deep learning and support vector machines \cite{frias2011agent, feng2020pmf, 10.1145/3381006}; the field is abundant with diverse methods, each offering a unique lens through which to understand human mobility. For example, gravity models and radiation models are founded on principles of physics. Gravity models, inspired by Newton's law of gravity, posit that the movement between two places is directly proportional to the size of their populations and inversely proportional to the distance between them \cite{erlander1990gravity}. Radiation models, on the other hand, are inspired by diffusion processes, assuming individuals move based on the opportunities in their immediate vicinity \cite{simini2012universal}. Both of these models have been traditionally employed to provide a macroscopic understanding of human mobility; however, recent studies have shown the shortcomings of these model in
reliable generation of human mobility flows at finer spatial temporal scales \cite{hsu2021limitations}.\\
\indent In recent years, machine learning methods have offered nuanced insights into individual and collective mobility patterns. For example, support vector machines (SVMs) offer a robust approach to classify data and make predictions, particularly for smaller, well-separated datasets \cite{frias2011agent}. Deep learning, a subset of machine learning leveraging neural networks with multiple layers (deep structures) to learn data representations and patterns, has also been used for relevant tasks \cite{bengio2013representation}. In human mobility prediction, deep learning techniques have shown the potential to model complex and nonlinear relationships, extract patterns from high-dimensional data, and forecast future states \cite{zhang2015spatiotemporal}. \\
\indent While the above methods have provided innovative tools for understanding and predicting human mobility, they have paid limited attention to the issues of fairness and equality of model outputs. The absence of fairness considerations in the existing human mobility prediction models could lead to pervasive inequalities in decisions and actions based on these models. The concept of fairness in the domain of human mobility prediction is of paramount importance and warrants serious attention. Fairness, in this context, signifies that the predictive models should demonstrate a uniform efficacy across various communities, regions, or demographic subgroups. The demand for such an impartial, consistent performance from these models springs from the need to ensure an equitable distribution of resources and development of infrastructure for all divisions of society, thereby upholding social justice and equity. \\
\indent The integration of machine learning models for human flow prediction into high-stakes decision-making processes has sparked concerns about possible biases that might inadvertently discriminate against specific groups. Such biases, highlighted in various research studies \cite{mukerjee2002multi,raghavan2020mitigating,berk2021fairness}, can lead to severe consequences if left unaddressed. For instance, a biased mobility model might under-perform and yield inaccurate predictions in rural regions, leading to misjudgment and under-investment in necessary infrastructure or services, thereby exacerbating regional disparities. The notion of fairness from the perspective of distributive justice entails an even-handed distribution of predictions, resources, or outcomes across various societal groups. Therefore, a fair human mobility prediction model would not over- or under-perform for a certain group based on their inherent characteristics, thus leading to equitable resource allocation and decision-making \cite{booth2021integrating}. Nevertheless, despite the criticality of fairness considerations in human mobility predictions, the current lack of fairness-aware models not only underscores a significant gap in effectiveness of the existing models, but it also directly undermines decision-making processes in aspects such as resource allocation, infrastructure development, and environmental planning, potentially propagating inequitable outcomes across various societal groups. \\
\indent To address this important gap, this study proposes a novel fairness-aware deep learning model (FairMobi-Net) for human flow prediction. The model is based on a specialized variant of the multi-linear perception that incorporates multiple sources of input features and adopts a source-oriented layer structure as its foundational architecture. The FairMobi-Net employs a three-stage approach to predict human mobility flow, by combining the outcomes of binary classification and numeric regression. Furthermore, the concept of fairness loss is introduced into the model's loss function to ensure fair outcomes across groups with variant income difference. \\
\indent To demonstrate the effectiveness of our proposed model, experiments are conducted using fine-grained real-world human mobility datasets in four U.S. cities. The results demonstrate that FairMobi-Net model can outperform the state-of-the-art  models (including the Deep Gravity Model), in terms of achieving more accurate as well as fairer human flow prediction across a variety of region pairs. When compared to the baseline models, FairMobi-Net is capable of producing predictions with comparable levels of accuracy consistently across different regions, demonstrating our model's advantage in maintaining a balance between high prediction accuracy and fairness. The interpretations of feature importance also reveal that certain features such as distance and road network structures as the important factors shaping the human flow across regions. 

\indent The model and findings will be particularly valuable to various academic disciplines and diverse stakeholder practitioners: (1) FairMobi-Net provides urban planners and transportation engineers with a novel method to fairly generate population flows for a variety of region pairs, regardless the differences of the median household incomes between regions; (2) The ability of the model to make a good trade-off between prediction accuracy and fairness ensures that the resultant mobility flows are equally accurate for equitable decision-making, such as infrastructure development and environmental planning; (3) the proposed Fairness Loss Function in the model provides a novel method for urban-computing researchers to improve the fairness performance of ML models in other urban applications; and (4) the evaluation of features that shape human mobility flows inform urban scientists and geographers about  influential factors, including social and built-environment features, that contribute to distribution of mobility flows across a city and subsequent outcomes, such as congestion, access, public health, air pollution, and economic activities. 



\section{Results}
\subsection{Dataset Collection and Experiment Settings} 
For experiments, we collect human mobility datasets from Spectus, which is a location intelligence and measurement platform collecting mobility data of anonymized devices. Data from about 15 M   active users are collected by Spectus in the United States. The previous studies have proven the high demographic representativeness of the Spectus dataset \cite{wang2019extracting,aleta2020modelling,nande2021effect}. Specifically, our human mobility datasets are collected at the census-tract level in four U.S. metropolitan areas 
\begin{itemize}
    \item 82,198 human mobility flows from the Atlanta Metropolitan Statistical Area (MSA).
    \item 507,994 human mobility flows from Harris County in Houston.
    \item 137,019 human mobility flows from King County in Seattle.
    \item 36,208 human mobility flows from Suffolk County in New York. 
\end{itemize}

\indent The data processing and modeling are performed using the NVIDIA RTX A6000 GPU. Based on the properties of the distributions, the learning rates of Atlanta MSA, Harris County, King County and Suffolk County are set 5e-4, 1e-3, 1e-3, 1e-3 respectively. We chose Adam \cite{kingma2014adam} as our optimizer with weight decay of 5e-4, and train each model with 1000 epochs.

\subsection{Human flow prediction}
\indent We conduct experiments for human flow prediction at the census-tract level within the above four areas. We adopt 60\% of the datasets for model training and another 20\% datasets as validation for hyperparameter fine tuning, and the remaining 20\% datasets for performance testing.
\begin{figure}[H]
\small
\centering
\includegraphics[width=12cm]{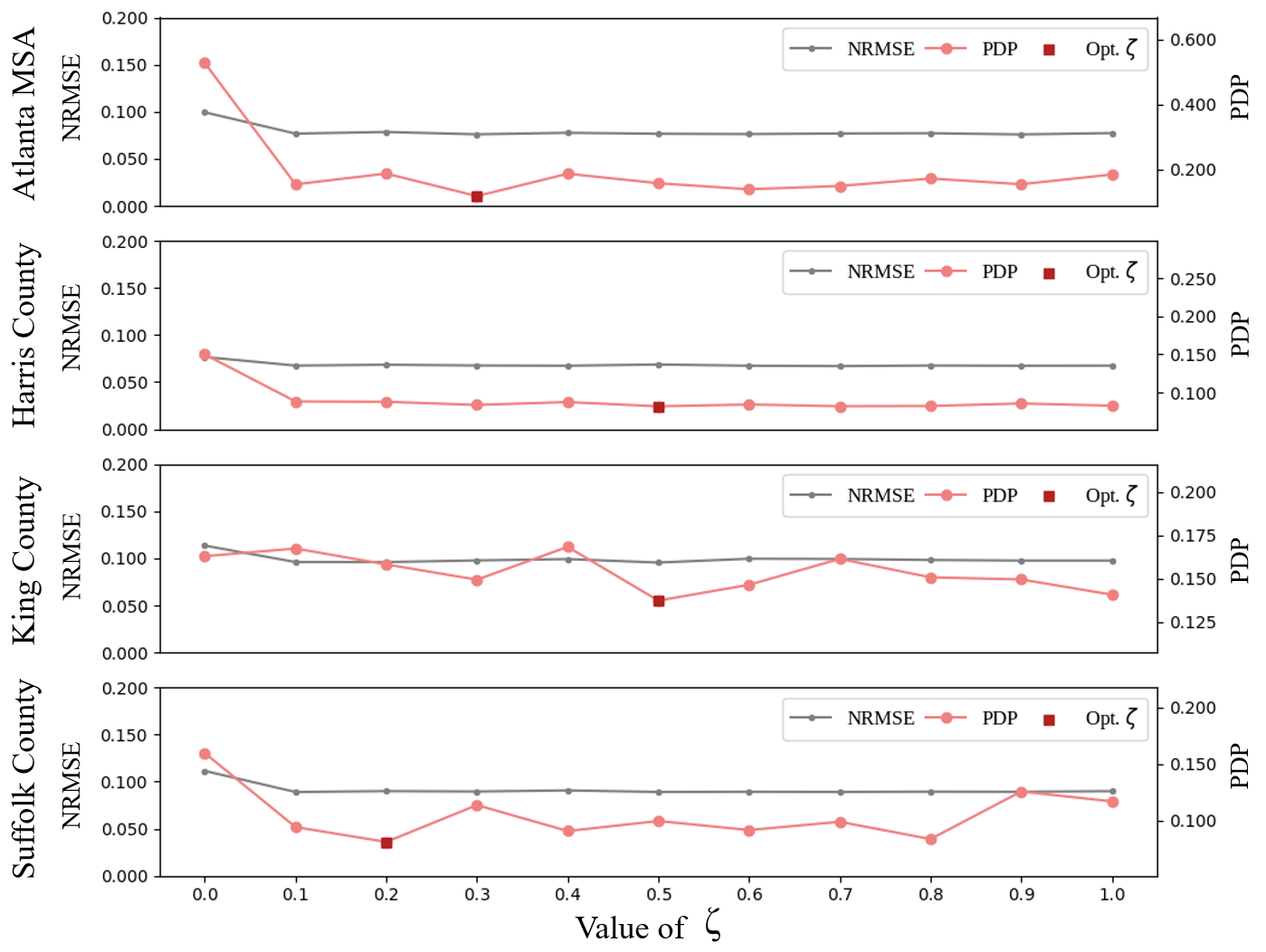}
\captionsetup{font=footnotesize}
\caption{\textbf{Selection of Lagrange multiplier coefficients selection.} For each region, we select the Lagrange multiplier coefficient that yields the lowest PDP for FairMobi-Net.}
\end{figure}
\indent The experiments aim to evaluate the effectiveness of the model in terms of the accuracy and fairness of human flow prediction. For comparison, we used the following models as the baselines:\\
•	Deep Gravity (DG): The latest state-of-the-art model for human flow prediction \cite{Simini2021}. \\
•	FairMobiNet-NoFL: A revised version of our proposed model FairMobi-Net, by removing the Fairness Loss from the FairMobi-Net’s loss function, to evaluate the effectiveness of introducing fairness loss  (which is a core contribution of this study)\\

\indent To determine the weight of fairness loss in the loss function, the Lagrange multiplier coefficient of the Fairness Loss needs to be defined by grid search. As shown in Figure 1, we iteratively test the Lagrange multiplier coefficient within the range of [0,1] with 0.1 intervals, and select the optimal value that yields the minimum PDP.\\

\indent Table 1 summarizes the performances of the models. It shows that the FairMobi-Net model can achieve accurate and fair predictions across different regions. For example, in Atlanta, FairMobi-Net has NRMSE = 0.076, 25.5\% lower than DG (NRMSE = 0.102), and 23.2\% lower than FairMobiNet-NoFL (NRMSE = 0.100), showing our model’s advantage in achieving more accurate human flow predictions than the baseline models. Figures 2 and 3, compare the ground-truth human flow with the predicted flow by respective models, which also shows the predicted human flow network by FairMobi-Net is more similar with the ground-truth human flow, through visual interpretation. Moreover, the DP achieved by the FairMobi-Net model is 0.118, significantly lower than DG (PDP = 0.607), indicating the performance by our model is fairer among different subpopulation communities than the DG model. The above results demonstrate the model outperforms the state-of-the-art baseline models, both in terms of human flow prediction accuracy and fairness performance. \\

\begin{table}
\centering
\captionsetup{font=footnotesize}
\caption{\textbf{Result table.} The only difference between MMLP and MMLP-FL is that the coefficient of MMLP-FL is 0.}
\begin{tabular}{|c|cccccc|}
\hline
Place & Model & NRMSE & PDP & Corr. & JSD & MAE \\
\hline
\multirow{3}{*}{Atlanta} & DG & 0.102 & 0.607 & 0.005 & 0.404 & 0.938 \\
 & FairMobiNet-NoFL & 0.100 & 0.528 & 0.121 & 0.391 & 0.900 \\
 & FairMobi-Net & \textbf{0.076} & \textbf{0.118} & \textbf{0.642} & \textbf{0.330} & \textbf{0.781} \\
\hline
\multirow{3}{*}{Harris} & DG & 0.102 & 0.272 & 0.781 & 0.382 & 1.997 \\
 & FairMobiNet-NoFL & 0.077 & 0.150 & 0.779 & 0.370 & 1.703 \\
 & FairMobi-Net & \textbf{0.069} & \textbf{0.082} & \textbf{0.801} & \textbf{0.353} & \textbf{1.610} \\
\hline
\multirow{3}{*}{King} & DG & 0.128 & 0.196 & 0.761 & 0.341 & 3.062 \\
 & FairMobiNet-NoFL & 0.113 & 0.163 & 0.756 & 0.320 & 2.710 \\
 & FairMobi-Net & \textbf{0.095} & \textbf{0.137} & \textbf{0.794} & \textbf{0.301} & \textbf{2.413} \\
\hline
\multirow{3}{*}{Suffolk} & DG & 0.131 & 0.198 & 0.782 & 0.342 & 3.185 \\
 & FairMobiNet-NoFL & 0.111 & 0.160 & 0.747 & 0.320 & 2.792 \\
 & FairMobi-Net & \textbf{0.090} & \textbf{0.081} & \textbf{0.819} & \textbf{0.284} & \textbf{2.394} \\
\hline
\end{tabular}

\label{tab:mytable}
\end{table}

\begin{figure}[H]
\small
\centering
\includegraphics[width=10cm]{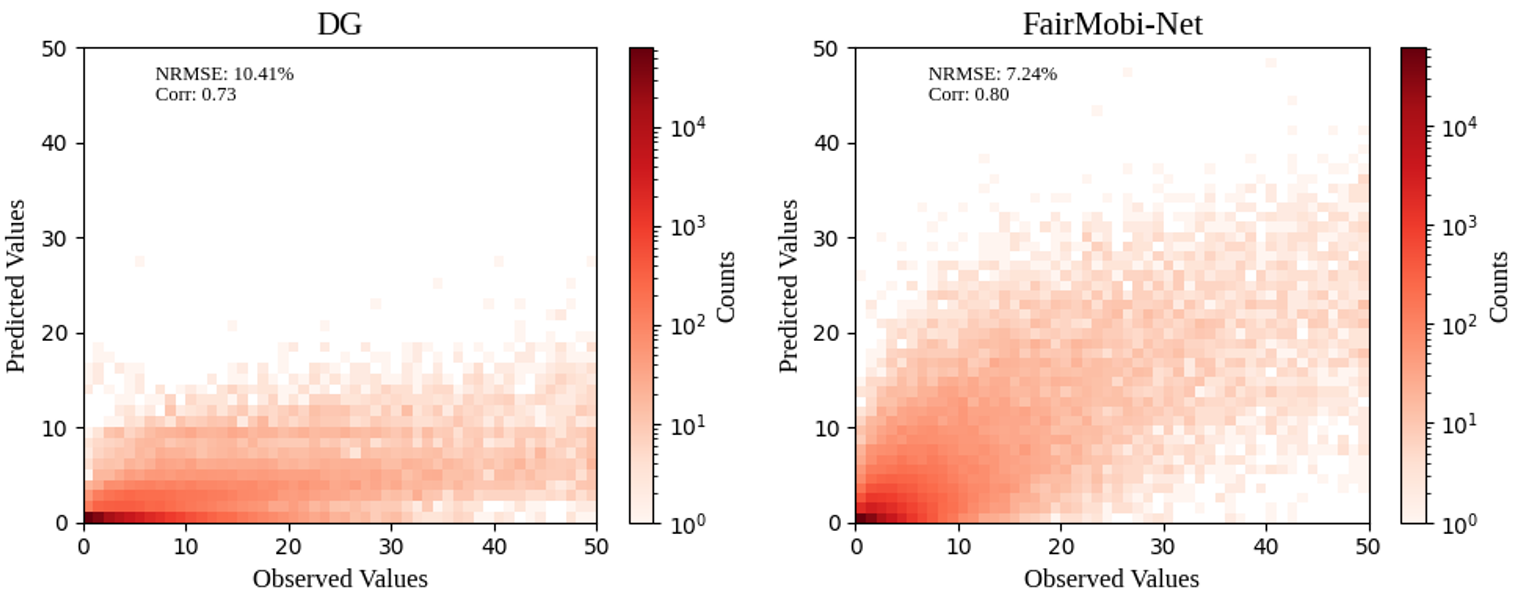}
\captionsetup{font=footnotesize}
\caption{\textbf{Correlation between the observed and predicted Flows} The results for the four study areas are aggregated. The comparison of the correlation between the observed and predicted flows by FairMobi-Net and DG shows that the predictions by FairMobi-Net have better consistency with the observed human flows than DG.}
\end{figure}

\begin{figure}[H]
\small
\centering
\makebox[\textwidth][c]{\includegraphics[width=\textwidth]{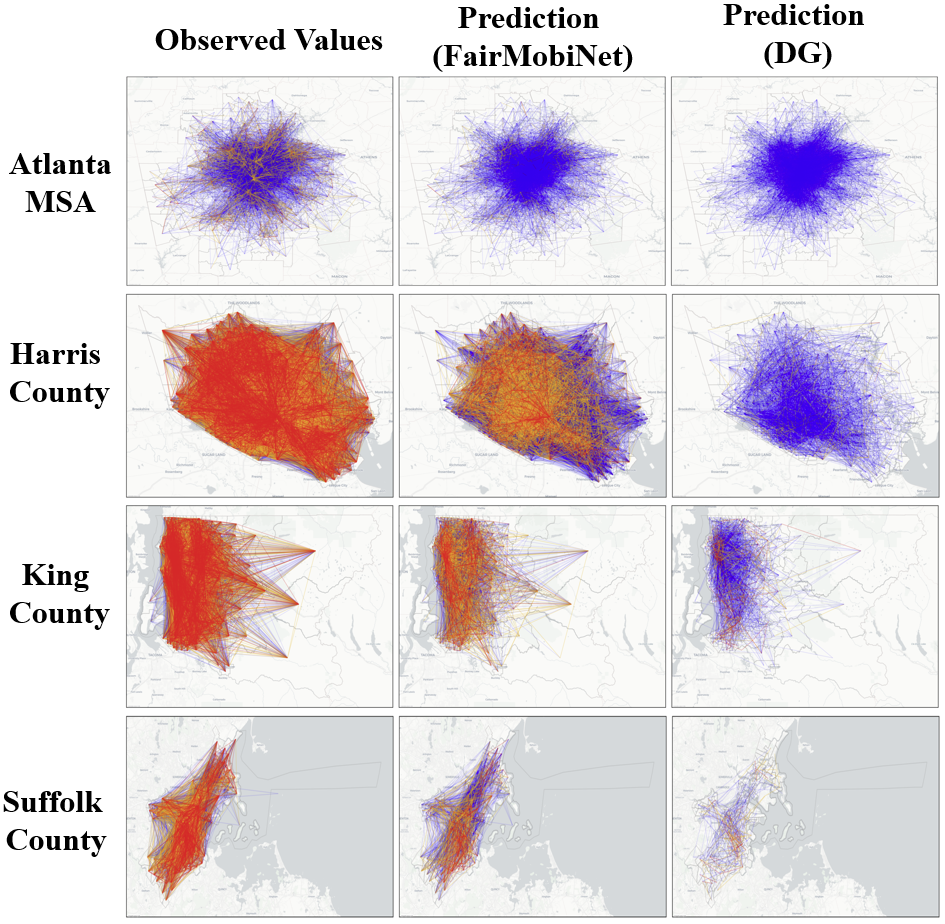}}%
\captionsetup{font=footnotesize}
\caption{\textbf{Spatial Comparison of Observed and Predicted Human Mobility Flows} The human mobility flows with different amounts are represented in different colors: blue for the human flow ranging from 1 to 3, orange for the human flow ranging from 4 to 9, and red for the human flow larger than 10. The visual interpretation clearly shows that the predicted human flows by our FairMobi-Net are more consistent with the observed flows, than DG}
\end{figure}

\indent Similar results can be obtained in Harris County and King County (Table 1): FairMobi-Net yields better accuracy (lower NRMSE, MAE, JSD and higher Corr.) and fairness (lower PDP) of prediction than other models. The performance of models are spatially depicted in Figure 4 shows that the models’ performance varies across the study regions, which may be due to the variations in regional characteristics, such as land use patterns, public facility availability, and road network configuration. This result suggests that the complex interplay of these diverse factors, which are inherently region-specific, may influence human flow dynamics, and subsequently, the predictive capacity of the model. In Suffolk County, the investigated models show similar prediction accuracy: NRMSE by our model is 0.090, 31.3\% lower than 0.131 of DG; while on the other hand, our model achieves significantly lower PDP (0.081 by FairMobi-Net), than that of DG (PDP= 0.198), proving the FairMobi-Net model’s particular advantage in achieving equal prediction across different communities. 
\begin{figure}[H]
\small
\centering
\makebox[\textwidth][c]{\includegraphics[width=\textwidth]{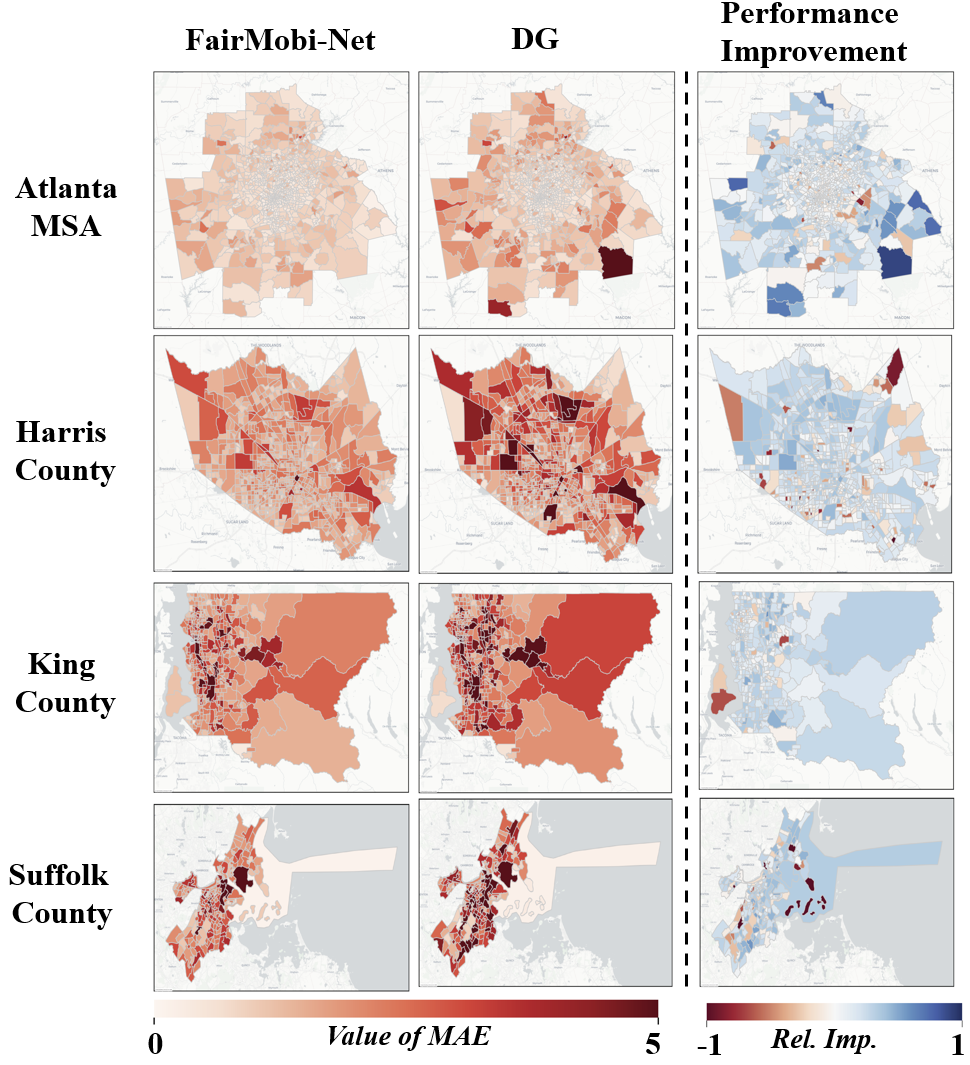}}%
\captionsetup{font=footnotesize}
\caption{\textbf{Spatial Comparison of Models' Accuracy} The average MAEs of FairMobi-Net and DP are calculated for respective census tracts. The third column plots the relative  improvement of FairMobi-Net over DG in terms of MAEs.}
\end{figure}

The results of further investigation of the models’ performance among different communities(Figure 5) shows that our model’s MAEs among different groups have less fluctuation than those of DG. The variances of MAE obtained by our model are 0.089 (Atlanta), 0.031 (Harris County), 0.029 (King County), 0.012 (Suffolk County), less than those by DG (0.122 (Atlanta), 0.043 (Harris County), 0.033 (King County), 0.016 (Suffolk County)). The results elucidated in Figure 5 and the corresponding variances underscore a remarkable consistency in the performance of our model across diverse communities. Our model not only demonstrates lower Mean Absolute Errors (MAEs), but also yields reduced fluctuations across regions when compared to the DG models. This consistency is manifested in significantly lower variances, notably in Atlanta, Harris County, King County, and Suffolk County. Such accurate and equitable performance across different groups substantiates the superior adaptability of our model, making it a more reliable and fair choice over DG models in various settings and demographics.\\

\begin{figure}[H]
\small
\centering
\includegraphics[width=10cm]{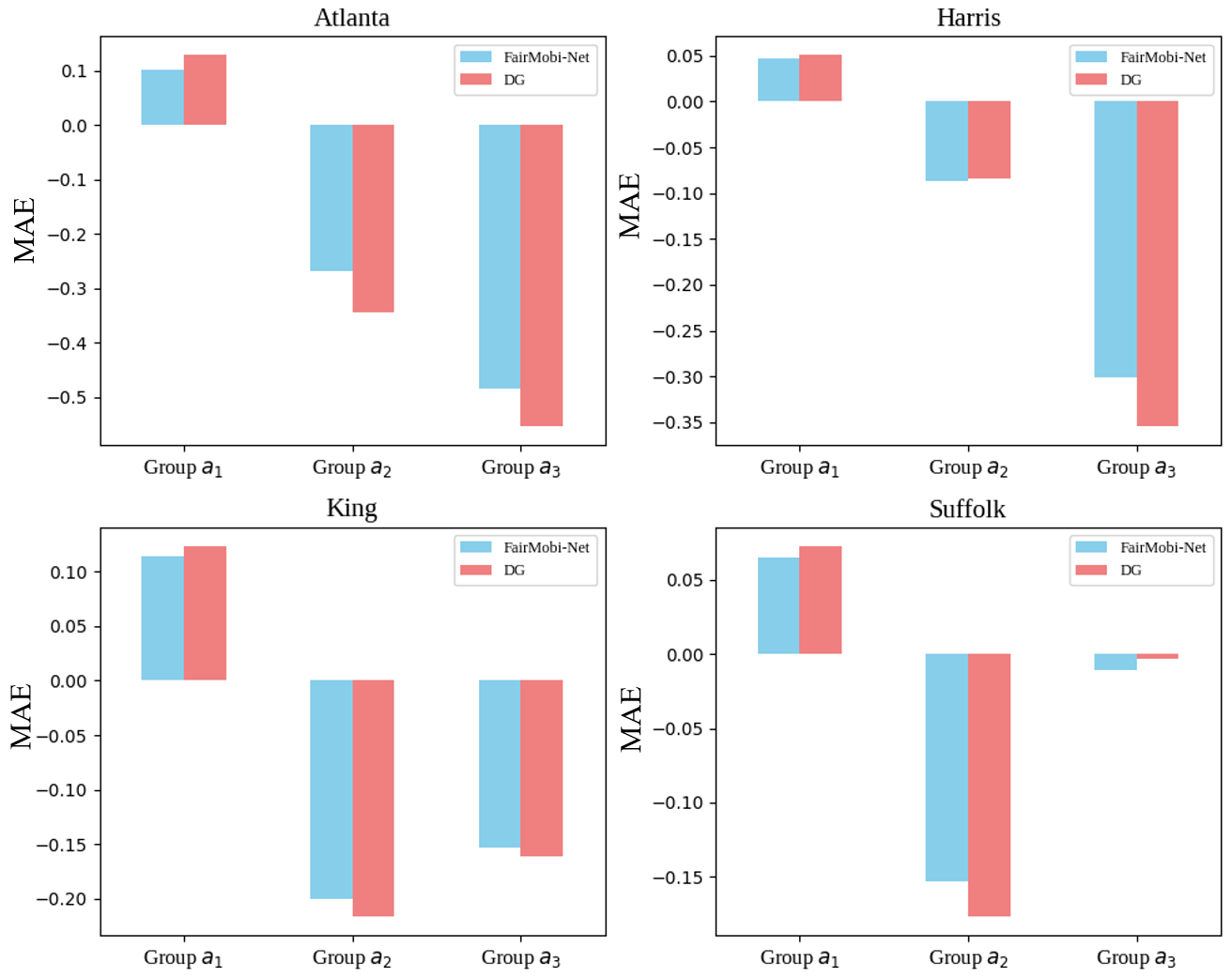}
\captionsetup{font=footnotesize}
\caption{\textbf{MAEs by FairMobi-Net and DG for Each Group} We plot the MAEs by FairMobi-Net and DG for census tract groups with various income difference (i.e.{a1, a2, a3}). The data indicates that, across all four study areas, the MAEs generated by FairMobi-Net exhibit fewer fluctuations than those produced by DG. This suggests that the proposed FairMobi-Net is capable of delivering predictions with a more uniform level of accuracy across varying groups.}
\end{figure}

\indent The introduction of Fairness Loss into the loss function leads to great improvement of models’ performance fairness. The most significant improvement is observed in Atlanta, where the PDP is improved from 0.528 (by FairMobiNet-NoFL) to 0.118 (by FairMobi-Net). A similar pattern is also consistent among other study regions. These marked improvements imply  that the introduction of Fairness Loss serves to reduce the disparity, ensuring that the model is more equitable in its predictive performance as well. Notably, the fairness element of the model aids in attenuating potential bias that may be inadvertently entrenched in the modeling process, thereby helping to accomplish more balanced and fair outcomes. Since this trend is not confined to Atlanta, it reflects a consistent pattern among other study regions as well, demonstrating the generalizability of the Fairness Loss addition across a diverse set of contexts and scenarios. The results attest to the robustness of the Fairness Loss strategy in facilitating model fairness, regardless of the location or area under consideration. A broad range of predictive models can benefit from these substantial advancements in performance equality by incorporating Fairness Loss. This innovation, therefore, opens up a promising avenue for advancing the development of predictive models for other urban computing applications that not only maintain high performance but also ensure the fair treatment of various subjects, areas, and groups they aim to predict for.

\subsection{Assessing the Influence of Urban Features on Human Flow Predictions}
The understanding and prediction of human flow between regions is a complex task that involves many factors. The interpretation of features importance in the FairMobi-Net model provides insights into how these factors shape the human flow. We used SHAP (SHapley Additive exPlanations) values to explain the output of our machine learning model and highlight the impact of various factors (Figure 6).\\

\begin{figure}[H]
\centering
\begin{minipage}{0.45\textwidth}
  \centering
  \includegraphics[width=\linewidth]{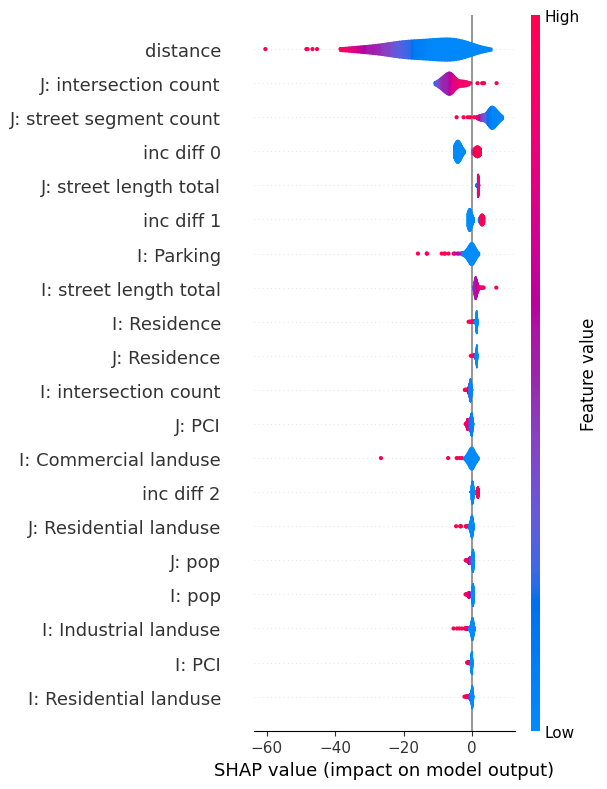}
  \captionsetup{font=footnotesize}
  \caption*{(a) Atlanta MSA}
  \label{fig:atlanta}
\end{minipage}\hfill
\begin{minipage}{0.45\textwidth}
  \centering
  \includegraphics[width=\linewidth]{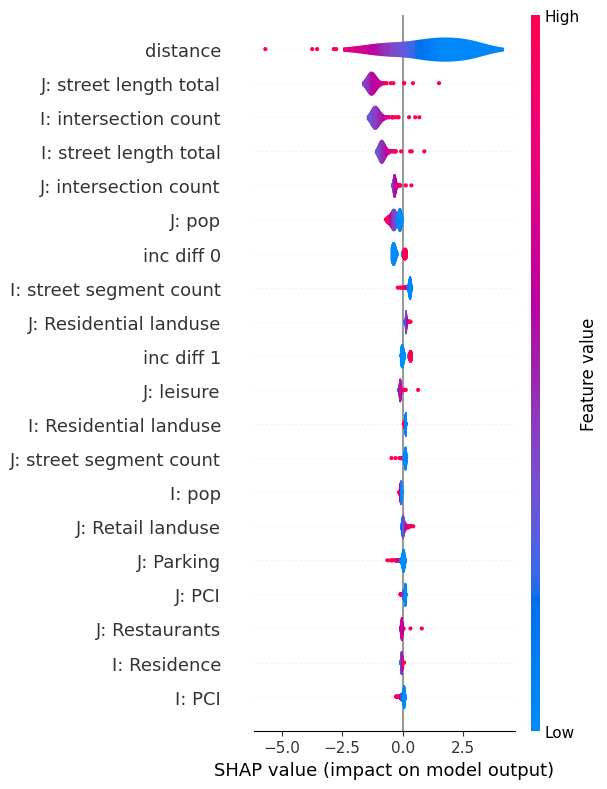}
  \captionsetup{font=footnotesize}
  \caption*{(b) Harris County}
  \label{fig:harris}
\end{minipage}
\begin{minipage}{0.45\textwidth}
  \centering
  \includegraphics[width=\linewidth]{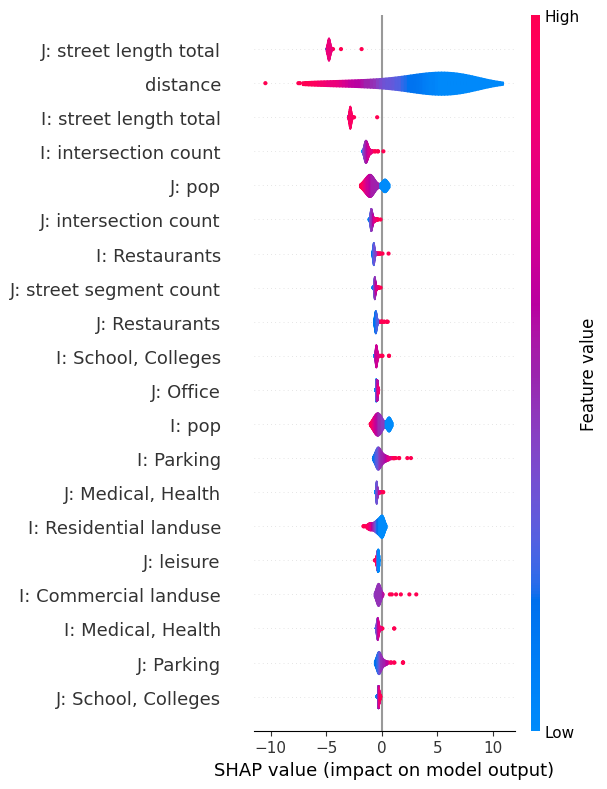}
  \captionsetup{font=footnotesize}
  \caption*{(c) King County}
  \label{fig:king}
\end{minipage}\hfill
\begin{minipage}{0.45\textwidth}
  \centering
  \includegraphics[width=\linewidth]{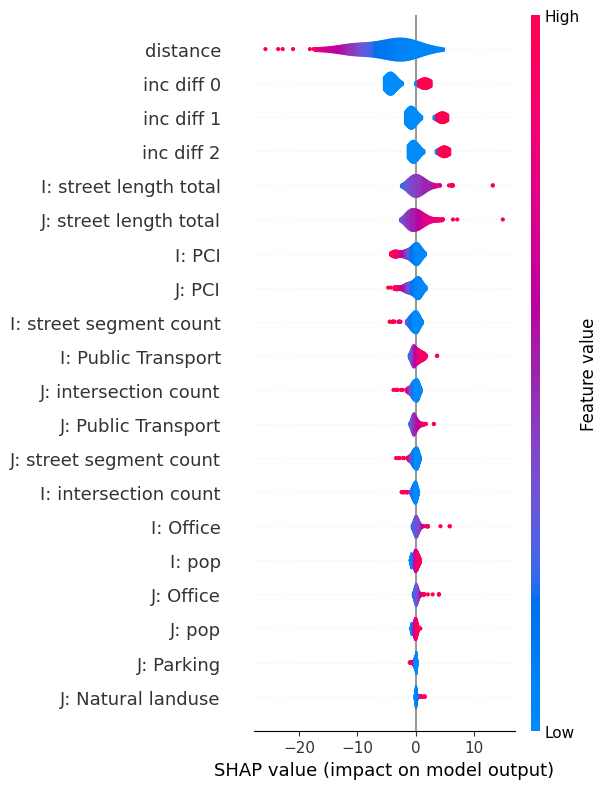}
  \captionsetup{font=footnotesize}
  \caption*{(d) Suffolk County}
  \label{fig:suffolk}
\end{minipage}
\caption{\textbf{SHAP Plots for Study Areas} On the graph, the x-axis represents the SHAP value, which signifies the degree to which a particular feature impacts the predictions. The color bar corresponds to the feature values themselves. Features prefixed with 'i' pertain to the origin census tract, while those prefixed with 'j' are derived from the destination census tract.}
\end{figure}

\indent Intuitively and consistent with the distance decay law of human mobility, distance emerged as the most influential factor in predicting human flow between regions, which is aligned with previous findings \cite{Simini2021}. This result aligns with the intuitive assumption that the larger the distance between regions, the less human flow between them. The high SHAP values of the distance factor, consistent across three out of four regions (i.e., Atlanta MSA, Harris County, Suffolk County), affirm the substantial role of the distance decay low of human mobility. This could be attributed to factors such as travel costs and time, which tend to increase with distance, thus discouraging high levels of human flow between distant regions. However, an interesting deviation from this general trend was observed in King County, where the total street length of the destination region was found to be the most influential factor, relegating distance to second place. This result could imply that the structural aspects of a region, represented here by street length, can significantly affect the human flow, perhaps even more so than the geographical distance. This might be because a larger total street length could indicate a more developed or dense urban environment, thus attracting more human flow.

In addition to distance and total street length, multiple other road network-related factors, such as total length of streets, total count of street segments, and total count of intersections, also showed considerable influence on the results. These factors might contribute to the fluidity of traffic and accessibility of different regions, thereby affecting the volume of human flow. This underlines the importance of considering not just geographical distance but also the quality and characteristics of infrastructure when predicting human flow. Conversely, other features such as land use, socioeconomic variables, and Points-of-Interest (POIs) generally exhibited minor influence on the human flow prediction. The direction of influence of these features also varied from region to region, indicating a possible interaction effect with local characteristics. For instance, certain types of land use or POIs might attract more human flow in one region due to cultural, demographic, or economic reasons.

In summary, while distance commonly plays a significant role in predicting human flow between regions, our findings underscore the importance of regional infrastructure, especially road networks, in shaping this flow. The influence of other factors, such as land use, socioeconomic variables, and POIs, appears to be more context-dependent, requiring further exploration for localized models. As we continue to refine our model, understanding these varying influences will help us increase its predictive accuracy and contribute more effectively to urban planning and policy making. This capability is particularly important for modeling mobility flows at finer spatial and temporal resolution. As the spatial and temporal resolution of human mobility prediction tasks increase, there is a need for consideration of additional features beyond the commonly known ones, such as distance, to achieve better accuracy and fairness performance.

\section{Problem Statement and Methodology}
\subsection{Problem Statement}
This study aims to predict the human flow between census tracts. Mathematically, this can be modeled as:
\begin{equation}
\begin{gathered}
y_{ij}=F(X_i, X_j, X_{i,j})
\end{gathered}
\end{equation}
where, $X_i$ represents the properties of the origin census tract $i$; $X_j$  represents the properties of the destination census tract $j$; $X_{i,j}$ is the communal features shared by census tracts $i$, $j$; $y_{ij}$ is the predicted human flow between census tracts $i$, $j$. Consequently, the goal of the model is to find an optimal mapping function $F$ from $(X_i,X_j,X_{i,j})$ to $y_{ij}$. The specifications for $F$ and features construction for $(X_i,X_j,X_{i,j})$ are given as described in the Methodology section, below.
\subsection{Methodology}
The model is designed to deliver highly accurate and fair predictions of human mobility flow. Our predictions are generated for pairs of census tracts, and by fairness, we mean that the accuracy of these predictions should be consistent across all tract pairs, irrespective of the disparity in income levels.
\subsubsection*{Proposed Model}
As shown in Figure 7, the input layer of our model is divided into three individual layer blocks. These blocks start with the local features $X_i$ from origin census tract $i$, the local features $X_j$ from destination census tract $j$ and the communal features $X_{i,j}$ between the census tracts respectively ($i, j\in {1, 2, …, N}$, where $N$ is the total number of census tracts within the region). Each block consists of a fully-connected layer, a Gaussian Error Linear Unit (GELU) \cite{hendrycks2016gaussian} activation function, a batch normalization layer and a dropout layer to introduce non-linearity. The features from the two census tract blocks are then combined by addition to match the topology of the features from the shared properties, and they are stacked together for further multiple block runs. Compared to the previous models, our model is designed for gravity-like patterns in the flow data and is more task-oriented and places greater emphasis on zero-inflated discrete distribution. Particularly, the prediction of human flow by our model is a three-stage process.

\begin{figure}[H]
\centering
\makebox[\textwidth][c]{\includegraphics[width=\textwidth]{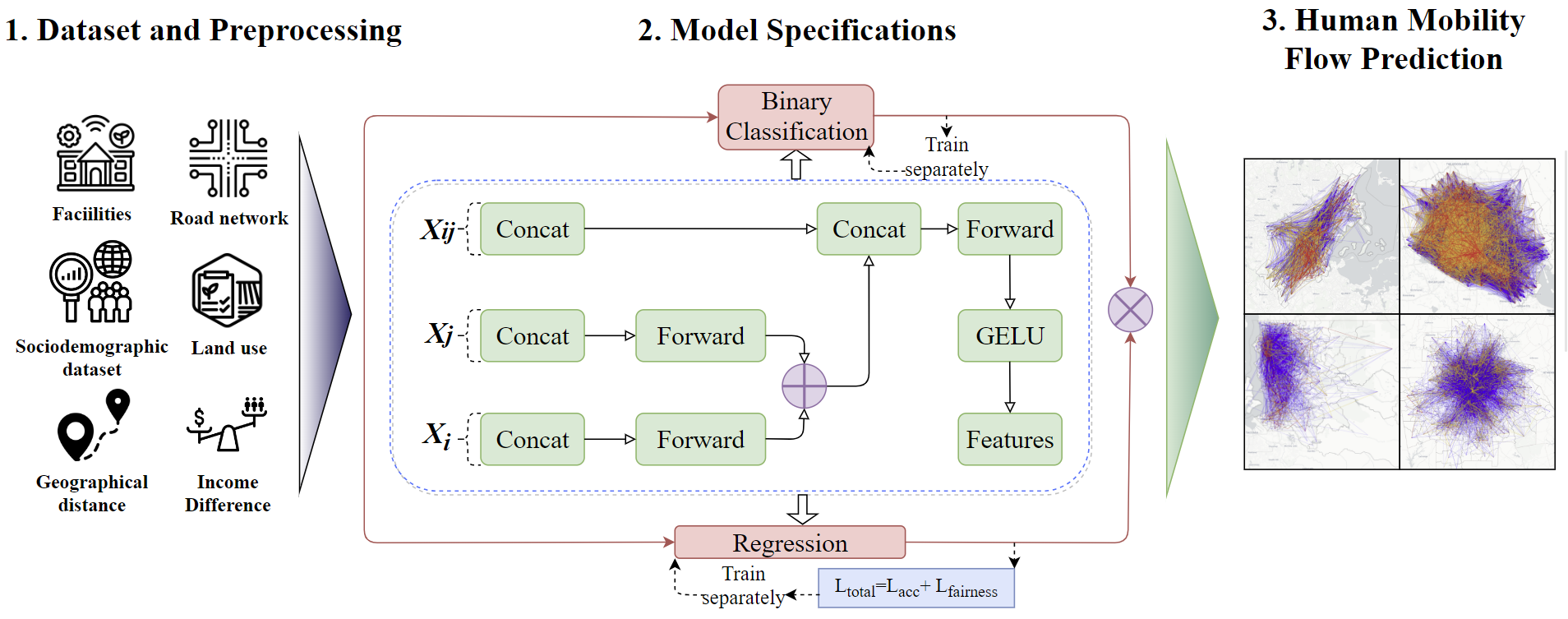}}
\caption{\textbf{The conceptual workflow of the study.} Multiple categories of datasets are collected to construct features for origin region and destination region. The proposed model introduces Fairness Loss into the loss function and utilizes a three-stage approach to predict the human mobility flows between regions.}
\end{figure}

\textbf{Stage One: Binary classification}\\ 
\indent In reality, the majority of the census tracts have no flow with other tracts, resulting in predominance of zero value in the training dataset. To handle this issue of imbalanced datasets, in this stage, our model is formulated as a binary classifier:
\begin{equation}
\begin{gathered}
    B_{ij}=M_1(X_i, X_j, X_{i, j})
\end{gathered}
\end{equation}
where, $B_{ij}$ is a binary value (0 or 1) to distinguish whether there is flow between the census tract. \\
\textbf{\indent Stage Two: Numeric regression}\\
\indent In this stage, our model is formulated as regression model: 
\begin{equation}
\begin{gathered}
    N_{ij}=M_2(X_i, X_j, X_{i, j})
\end{gathered}
\end{equation}
where $N_{ij}$ is a numeric value indicating the possible inter-tract human.\\
\textbf{\indent Stage Three: Human flow prediction}\\
\indent In this stage, the output of previous stages is combined to give the final prediction of human flow:
\begin{equation}
\begin{gathered}
    y_{ij}=F(X_i, X_j, X_{i, j})=B_{ij}\cdot N_{ij}
\end{gathered}
\end{equation}
where $y_{ij}$ is the final predicted human flow between tract $i$ and $j$.
\subsubsection*{Loss Specifications}
The key motivation for our study is to achieve predictions with similar accuracy across the different census tract groups. Consequently, we introduce the concept of Fairness Loss into the loss function of our model:
\begin{equation}
\begin{gathered}
L_{tot} = \Bar{l}+\zeta\cdot L_{fairness}
\end{gathered}
\end{equation}
where, $\Bar{l}$ is the loss function characterizing the accuracy of human flow prediction; $L_{fairness}$ is the introduced loss to characterizing the fairness of model prediction across groups; $\zeta$ is a Lagrangian multiplier that determines the weight of $L_{fairness}$.\\ 
\indent In our model, accuracy loss $\Bar{l}$ is defined as the mean absolute error (MAE) of the prediction
\begin{equation}
\begin{gathered}
\Bar{l}=\frac{1}{N}\sum^N_{i=1}\sum^N_{j=1}|y_{ij}-\Bar{y_{ij}}|
\end{gathered}
\end{equation}
where, $\Bar{y_{ij}}$ is the predicted human flow (see Equation 4), $y_{ij}$ is the ground-truth observed human flow.\\
\indent Correspondingly, the aim of achieving similar accuracy across different groups is to explore the optimal solutions of the following optimization problem:
\begin{equation}
\begin{gathered}
\min: \quad \scalebox{0.8}{$\Bar{l}$}\\
\textrm{s.t.} \quad \scalebox{0.8}{$P(\Bar{l}-\tau\leq l \leq \Bar{l}+\tau|a_1 )=P(\Bar{l}-\tau\leq l \leq \Bar{l}+\tau|a_2 )$}\\
\scalebox{0.8}{$...$} \\
\scalebox{0.8}{$\quad P(\Bar{l}-\tau\leq l \leq \Bar{l}+\tau|a_1 )=P(\Bar{l}-\tau\leq l \leq \Bar{l}+\tau|a_n )$}\\
\scalebox{0.8}{$...$} \\
\scalebox{0.8}{$\quad P(\Bar{l}-\tau\leq l \leq \Bar{l}+\tau|a_n )=P(\Bar{l}-\tau\leq l \leq \Bar{l}+\tau|a_{n-1})$}\\
\scalebox{0.8}{$\forall a_j\in A,\ j\in\{1, 2, ..., N_A\}$} \\
\end{gathered}
\end{equation}
where, $a_i$ represents a group of human mobility predictions between certain areas (details of the grouping are given in the next section); the number of human flow predictions related in $a_i$ is represented as $N_{a_i}$; $A$ is the total collection of groups of census tracts; $l$ is the accuracy loss array of group $a_i$;  $P(\cdot|a_i)$ is the probability that the error of each prediction in $a_i$ falls in the range delimited by $\Bar{l}$ and a threshold $\tau$. The above formula illustrates the searching for global optima that satisfy the following condition: for each group $a_i$, the accuracy loss should have similar distribution (characterized as a range determined by $\Bar{l}$ and $\tau$) as any other group. \\
\indent To solve this problem by modeling, we reformulate the optimization problem using a Lagrangian. Considering that the difference in population of each group in each area is imbalanced, it is advised to multiply a weight $w_{p,q}=\frac{\sum_{a_i\in A}N_{a_i}}{N_{a_p}+N_{a_q}}$ to amplify the impact of less populous groups. Hence, the optimization problem depicted in equation (7) can be transformed into a Lagrangian:
\begin{equation}
\begin{gathered}
\min \quad\scalebox{0.8}{$ L(y, \zeta)=l+\sum_{p,q\in A,p\neq q}\zeta_{p,q}w_{p,q}|P(\Bar{l}-\tau\leq l \leq \Bar{l}+\tau|a_p )-P(\Bar{l}-\tau\leq l \leq \Bar{l}+\tau|a_q)|$}\\
\textrm{$\Rightarrow$} \quad \scalebox{0.8}{$L(y, \zeta)=l+\zeta\cdot\sum_{p,q\in A,p\neq q}w_{p,q}|P(\Bar{l}-\tau\leq l \leq \Bar{l}+\tau|a_p )-P(\Bar{l}-\tau\leq l \leq \Bar{l}+\tau|a_q)|$}\\
\textrm{$\Rightarrow$} \quad \scalebox{0.8}{$L(y, \zeta)=l+\zeta\cdot PDP$}\\
\textrm \quad \scalebox{0.8}{$PDP=\sum_{p,q\in A,p\neq q}w_{p,q}|P(\Bar{l}-\tau\leq l \leq \Bar{l}+\tau|a_p )-P(\Bar{l}-\tau\leq l \leq \Bar{l}+\tau|a_q)|=L_{fairness}$}
\end{gathered}
\end{equation}
To simplify the process of exploring the optimal coefficient that achieves optimal fairness performance with acceptable loss, we estimate each equity Lagrangian multiplier as a constant value and define the summation part as Proportional Demographic Parity (PDP). In this specific case study, we interpret PDP as representing the loss in fairness. PDP focuses on reducing the distance between each pair of groups based on the proportion of prediction losses falling within the domain defined by the overall loss in modeling, and we use it as a key metric for evaluating the fairness performance of our model hereafter in the study.

\subsubsection*{Feature Construction and Prediction Grouping}
As explained in Equation (1), multi-dimensional feature $X_i$, $X_j$, $X_{i,j}$ are constructed to present the properties of the origin and destination census tracts. The specific features are constructed as below:
\begin{itemize}
    \item $X_i$, $X_j$:
    \begin{itemize}
        \item Public facilities (nine features): total count of Points-of-interest (POIs) and buildings relevant to restaurant, school \& college, public transport, office, leisure, medical \& health, residence, parking and retail.
            \item Land-use (six features): total area (in $km^2$) for commercial, construction, industrial, residential, retail and natural land-use classes.
            \item Road network (three features): total length (in $m$) of streets, total count of street segments and total count of intersections.
        \item Census statistics (two features): population and per capita income (in \$). 
    \end{itemize}
    \item $X_{i,j}$
    \begin{itemize}
        \item Euclidean distance between two census tracts (in feet).
        \item Protected attribute (three features): the one-hot encoded income difference between census tracts. We rank the absolute difference of median household income between origin and destination census tracts in ascending order, and define groups of human mobility prediction as:
        \begin{itemize}
            \item \textbf{Group} $a_1$—the top 20\% ranking, as the group of human mobility predictions between census tracts with low income difference.
            \item  \textbf{Group} $a_2$—20\% to 50\% ranking, as the group of human mobility predictions between census tracts with medium income difference. 
            \item \textbf{Group} $a_3$—The remaining 50\%, as the group of human mobility predictions between census tracts with high group income difference. \\
        \end{itemize}
    \end{itemize}
\end{itemize}
\indent The total number of features is 44, of which 20 belong to the origin census tract, 20 features are from the destination census tract, and the remaining 4 are communal features. \\
\indent \textbf{The aim of our model} is hence to achieve accurate as well as fair prediction for census tract groups with income differences (i.e., $A=\{{a_1,a_2,a_3}\}$ in Equation 7, and the achieved accuracy for human flow prediction should be approximately consistent across  groups $a_1$, $a_2$ and $a_3$).

\subsection*{Evaluation metrics}
We adopted the following measurements to evaluate the performance of our model. For accuracy evaluation, four measurements are adopted: Normalized Root Mean Square Error (NRMSE) and mean absolute error (MAE), both to measure the quantitative deviations of the predicted human flow from observed values. Lower values of the measurements denote more accurate predictions \cite{tadj2014improving}. Pearson's Correlation (Corr.) assesses the inter-dependency between input features and the target variable. A higher value Corr. indicates stronger correlation relationships \cite{godfrey1980correlation}. Jensen-Shannon Divergence (JSD) measures the similarity between the distribution of predicted values and observed values. A lower JSD value implies more similar distributions \cite{fuglede2004jensen}. For fairness evaluation, we mainly aim to evaluate the equality of model performance among different income difference groups, hence, the measurement PDP is used (Equation 8). A lower value of PDP implies a more equitable prediction outcome.

\section{Closing Remarks}
The attainment of accurate and fair human flow predictions across diverse socioeconomic subgroups is important for promoting social justice and for aiding in decision making for equitable resource allocation and urban development. In this study, we introduce a novel fairness-aware deep learning model, FairMobi-Net, designed for human flow prediction. Employing a three-stage approach, this proposed model incorporates the concept of Fairness Loss and employs a three-stage approach to predict human flow between pairs of regions. Experimental results using real-world datasets from four U.S metropolitan areas indicate that our model  delivers human mobility flow predictions with high accuracy across regions. Compared with the baseline models, FairMobi-Net yields comparable levels of accuracy consistently across different regions, showcasing our model's strength in ensuring both good accuracy and fairness in human mobility prediction.\\
\indent The model and outcomes of this study have multiple important contributions. From a theoretical perspective, the FairMobi-Net model provides a unique fairness-aware approach to human mobility prediction by introducing a novel fairness loss component in the deep learning model. The revised loss function ensures the model achieves a good trade-off between prediction accuracy and fairness. The novel fairness loss provides a new approach for future fairness-oriented model design for other urban computing applications as well. From a practical standpoint, the model offers a novel tool for urban planners, transportation engineers, and environmental managers involved in decision-making processes, such as infrastructure and environmental planning, to better predict future patterns of human mobility flows based on anticipated changes in the land use and development patterns with the equity in mind. The model helps ensure that the resultant mobility flows are accurate and fair for all communities, which promotes equitable resource allocation and infrastructure development and, ultimately, social justice.\\
\indent Moving forward, future research could aim to extend the application of the FairMobi-Net model to different geographic locations and settings, exploring how the consideration of fairness improves the machine learning models in other urban computing applications (such as air-pollution prediction). Another intriguing avenue to explore is how to fine-tune the FairMobi-Net model to take into account other demographic factors that might influence human mobility flows, such as age, gender, or occupation as sensitive attributes. Finally, with the growing application of machine learning models in urban studies, integrating the FairMobi-Net model with other predictive models used in sectors like public health, economics, or environmental studies could also provide holistic insights into human mobility behavior and its implications on other urban outcomes (such as air pollution, prevalence of diseases, and economic activity) at a larger scale.

\section*{Data Availability and Acknowledgement }

This material is based in part upon work supported by the National Science Foundation under Grant
CMMI-1846069 (CAREER). Any opinions, findings, conclusions, or recommendations expressed in this material are those of the authors and do not necessarily reflect the views of the National Science Foundation.

\bibliography{ref}

\end{document}